\documentclass[conference]{IEEEtran}
\IEEEoverridecommandlockouts
\usepackage{cite}
\usepackage{amsmath,amssymb,amsfonts}
\usepackage{algorithmic}
\usepackage{graphicx}
\graphicspath{ {./images/}}
\usepackage{textcomp}
\usepackage{xcolor}
\usepackage{pgfplots}
\usepackage{subcaption}
\usepackage{tikz}
\usepackage{balance}
\usetikzlibrary{shapes.geometric, arrows, positioning, calc}
\pgfplotsset{width=9cm,compat=1.9}
\usepackage[bottom]{footmisc}
\usepackage{url}
\usepackage[hidelinks]{hyperref}
\def\BibTeX{{\rm B\kern-.05em{\sc i\kern-.025em b}\kern-.08em
    T\kern-.1667em\lower.7ex\hbox{E}\kern-.125emX}}
\begin{document}

\title{Classification of Cattle Behaviour and Detection of Heat(Estrus) using Sensor Data
\\
\thanks{Centre for Development of Advanced Computing}
}

\author{\IEEEauthorblockN{Druva Dhakshinamoorthy \textsuperscript{$\dagger$}}
\IEEEauthorblockA{\textit{BITS Pilani, K. K. Birla Goa Campus} \\
Goa, India \\
f20220131@goa.bits-pilani.ac.in}
\and
\IEEEauthorblockN{Avikshit Jha\textsuperscript{$\dagger$}}
\IEEEauthorblockA{\textit{BITS Pilani, Pilani Campus} \\
Pilani, India \\
f20220479@pilani.bits-pilani.ac.in}
\and
\IEEEauthorblockN{Sabyasachi Majumdar}
\IEEEauthorblockA{\textit{Centre for Development of Advanced Computing} \\
Kolkata, India \\
sabyasachi.majumdar@cdac.in}
\and
\IEEEauthorblockN{Devdulal Ghosh}
\IEEEauthorblockA{\textit{Centre for Development of Advanced Computing} \\
Kolkata, India \\
devdulal.ghosh@cdac.in,}
\and
\IEEEauthorblockN{Ranjita Chakraborty}
\IEEEauthorblockA{\textit{Centre for Development of Advanced Computing} \\
Kolkata, India\\
rinaranjita@gmail.com}
\and
\IEEEauthorblockN{Dr. Hena Ray}
\IEEEauthorblockA{\textit{Centre for Development of Advanced Computing} \\
Kolkata, India\\
hena.roy@cdac.in}
}

\maketitle
\def\thefootnote{\textsuperscript{$\dagger$}}\footnotetext{These authors contributed equally to this work and are co-first authors.}

\begin{abstract}
This paper presents a novel approach to cattle behavior detection and detection of Estrus(heat) periods using sensor data and machine learning techniques. By employing an affordable Bluetooth-based monitoring system, we aim to enhance livestock management and improve animal welfare. Our system uses a neck collar equipped with accelerometer and gyroscope sensors to gather real-time behavioral data from the cattle and syncing it to the cloud. We evaluated three different machine learning algorithms—Support Vector Machines (SVM), Random Forests (RF), and Convolutional Neural Networks (CNN)—to classify various cattle behaviors accurately. Additionally, we implemented a Long Short-Term Memory (LSTM) network to detect estrus by analyzing the patterns in these classified behaviors.
\end{abstract}


\section{Introduction}
The agriculture industry is experiencing a significant transformation with the advent of precision farming techniques. Among these, smart livestock monitoring has emerged as a crucial area of focus, particularly in the dairy and cattle farming sectors. This paper presents a novel approach to cattle behavior detection and prediction using sensor data and machine learning algorithms, aiming to revolutionize how farmers monitor and manage their herds.
Cattle health monitoring is critical for ensuring animal welfare and optimizing farm performance. Traditional observation methods are often time-consuming, labor-intensive, and may miss subtle behavioral changes that could indicate health issues or critical physiological states. The need for more efficient, accurate, and real-time monitoring solutions has driven the development of technological interventions in this field.
Our project, conducted at the Centre for Development of Advanced Computing (CDAC) in Kolkata, aims to analyze key cattle behaviors such as feeding, rumination, and lying patterns. We utilize accelerometer and gyroscope sensors embedded in a low-cost, Bluetooth-based device mounted on a neck collar worn by the cattle. This system collects real-time behavioral data and syncs it the Influx DB. These behaviors are important indicators of cattle health, well-being, and reproductive status. \cite{behaviour_health} \cite{REITH2018398}.
The primary motivation behind this research is to enhance livestock management through advanced monitoring techniques. By providing farmers with early warning signs of potential health issues and insights into the estrus cycles of cattle, we aim to enable more timely interventions, reduce treatment costs, and improve overall herd management. This is particularly crucial in large-scale farming operations where manual inspection of individual animals becomes increasingly challenging.
Our approach combines sensor technology with machine learning algorithms to transform raw data into actionable insights. This paper details our methodology, including data collection, preprocessing techniques, and the evaluation of various machine-learning models. We also discuss the challenges encountered during the project and outline future directions for this research.
By developing an accessible and efficient tool for precision livestock farming, we aim to contribute to improved animal welfare, increased farm productivity, and ultimately, a more sustainable and technologically advanced agricultural sector.

\section{Background}

The advent of sensor technology and machine learning algorithms has opened up new possibilities for more efficient, accurate, and real-time monitoring of cattle. Wearable sensors, such as neck collars equipped with accelerometers and gyroscopes, can collect continuous data on cattle behavior. This data, when properly analyzed, can provide valuable insights into various aspects of cattle health and welfare. \cite{Miche}

Key behaviors such as feeding, rumination, and lying patterns are important indicators of cattle health, well-being, and reproductive status. By accurately detecting and classifying these behaviors, farmers can gain early warning signs of potential health issues and insights into the estrus cycles of cattle. This enables more timely interventions, reduces treatment costs, and improves overall herd management. \cite{chang}

The challenge lies in transforming the raw sensor data into actionable insights. This is where machine learning algorithms come into play, offering the ability to process large amounts of data and identify patterns that might be difficult for humans to detect. Various machine learning approaches, including Support Vector Machines (SVM), Random Forests (RF), and Convolutional Neural Networks (CNN), have been explored for behavior classification. \cite{barwick} \cite{chang}

For more complex tasks such as estrus detection, which requires understanding of cow behaviour and time period at which it occurs, more advanced techniques like Long Short-Term Memory (LSTM) networks have shown promise \cite{nan_lstm}. These approaches aim to provide farmers with tools for precision livestock farming, contributing to improved animal welfare, increased farm productivity, and a more sustainable agricultural sector.
\section{Related Work}
\subsection{Behaviour Classification}
The application of sensor technology and machine learning algorithms for monitoring cattle behavior, particularly rumination, has gained significant attention in recent years. Several studies have explored various approaches to predict and classify rumination patterns using accelerometer data.

Michie et al. used Fourier transform analysis of accelerometer data to predict rumination, demonstrating a novel approach to behavior detection \cite{Miche}.
Chang et al. compared various methods for predicting rumination in cattle using tri-axial accelerometer ear-tags. \cite{chang}
Barwick et al. implemented a machine learning model using triaxial accelerometer data on cattle, further advancing the application of this technology in livestock management. \cite{barwick}
These studies collectively indicate that accelerometer-based sensors, especially ear-tags, show promise for monitoring various ruminant behaviors. The research demonstrates the versatility of accelerometer data, from Fourier transform analysis to machine learning applications.

\subsection{Estrus Detection}
Shahriar et al.~\cite{shahriar_unsup} proposed an unsupervised method using accelerometer data:
\begin{itemize}
    \item Segmented time series data into windows
    \item Extracted features (standard deviation, amplitude, energy, FFT) \cite{fft}
    \item 
    Applied K-means clustering for activity intensity levels
    \end{itemize}
    They clustered high and medium activity levels, evaluated Activity Level \(\gamma_t\) which is [count(High)*0.9 + count(Medium)*0.1], and finally identified anomalies in \(\delta_t\) which is a parameter comparing \(\gamma_t\) with the previous values of \(\gamma_t\), to flag heat conditions. The expression was as follows:

    \[
\delta_t = \frac{3\gamma_t - (\gamma_{t-72} + \gamma_{t-48} + \gamma_{t-24})}{\gamma_t + \gamma_{t-72} + \gamma_{t-48} + \gamma_{t-24}}
\]

where:
\begin{itemize}
    \item \(\delta_t\) is the high activity index for time \(t\) (hourly).
    \item \(\gamma_t\) is the activity level at time \(t\).
    \item \(\gamma_{t-N}\) is the activity level N hours (N/24 days) before time \(t\).

\end{itemize}
 We found that the high activity level data correlated to the others activity data labeled by the classification model, medium activity to the feeding data, and lower activity to the rumination and lying data.

\subsection{LSTM-based Prediction}
Nan Ma et al.~\cite{nan_lstm} developed a deep learning approach that utilizes NB-IoT for data collection and employs LSTM and CNN for estrus prediction. This method outperformed existing algorithms in accuracy and efficiency. However, it uses lot more sensor data to derive the required information for the LSTM network.

\subsection{Comparison}
While Shahriar et al.'s method offers a straightforward, unsupervised approach, Nan Ma et al.'s solution provides a more sophisticated, predictive model. The latter's use of LSTM networks allows for better access to behavioral patterns, improving estrus detection accuracy. Our approach utilizes the behavior patterns categorized by the Behaviour Classification model and utilizes this labeled data to train an LSTM model, utilizing anomaly detection \cite{anomaly_detec} to detect the Estrus condition with fewer sensor data and similar accuracy.

% Remember to add these citations to your bibliography
% \bibitem{shahriar} Shahriar et al., [Title], [Journal], [Year]
% \bibitem{nanma} Nan Ma et al., [Title], [Journal], [Year]
\section{Behaviour Classification}
\subsection{Analysis of ML Models}
\begin{figure}[h]
\centering
\begin{tikzpicture}
    \begin{axis}[
        width=0.49\textwidth,
        x tick label style={/pgf/number format/1000 sep=},
        symbolic x coords={1,5,15},
        ylabel={Training times (minutes)},
        xlabel={Dataset size (days)},
        enlargelimits=0.15,
        legend style={at={(0.5,-0.25)},
        anchor=north,legend columns=-1},
        ybar,
        bar width=10pt,
        xtick=data,
        ]
    \addplot 
	coordinates {(1,3.22)(5,20)(15,0)};
    \addplot 
	coordinates {(1,0.05)(5,0.5)(15, 1.7)};
    \addplot 
	coordinates {(1,0.8)(5,2.1)(15,3.8)};
    \legend{SVM,RF,CNN}
    \end{axis}
\end{tikzpicture}
\caption{Training time of Models vs Dataset Size\textsuperscript{*}}
\label{fig:training_time}
\end{figure}
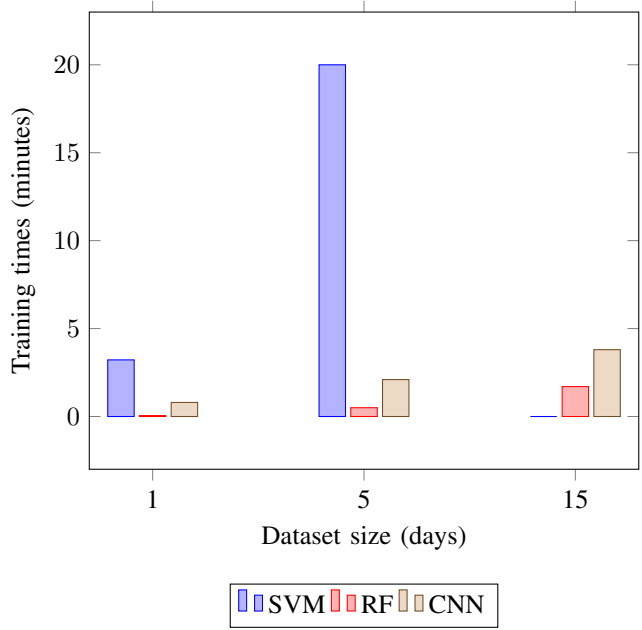

Three machine learning algorithms were evaluated for their effectiveness in classifying cattle behavior: Support Vector Machines (SVM), Random Forests (RF) \cite{random_for}, and Convolutional Neural Networks (CNN)    \cite{CNN}.
SVMs, while memory-efficient and effective in high-dimensional spaces, showed adequate performance but were generally outperformed by the other algorithms. SVMs faced challenges with longer training times on large datasets as shown in Fig \ref{fig:training_time} , and was hence not a viable algorithm to be used in a production environment.
Random Forests demonstrated superior performance in predicting cattle behavior. Its ability to handle large datasets with multiple features aligned well with the sensor data collected. RF also provided feature importance rankings, helping identify critical sensor measurements for behavior prediction. Its resistance to overfitting and capability to work with both categorical and continuous variables proved advantageous\cite{random_for}.

\begin{figure}[t]
\begin{subfigure}[b]{0.235\textwidth}
        \includegraphics[width=\textwidth]{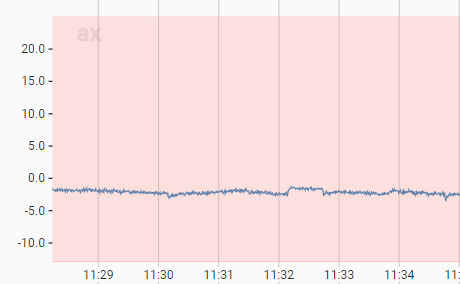}
        \caption{ax Rumination Pattern}
        \label{fig:ax_rumin}
    \end{subfigure}
    \begin{subfigure}[b]{0.235\textwidth}
        \includegraphics[width=\textwidth]{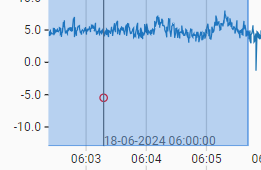}
        \caption{Feeding Pattern}
        \label{fig:feeding}
    \end{subfigure}
    \begin{subfigure}[b]{0.235\textwidth}
        \includegraphics[width=\textwidth]{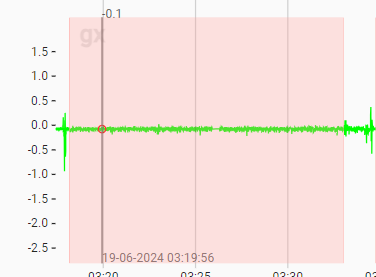}
        \caption{gx Rumination Pattern}
        \label{fig:gx_rumin}
    \end{subfigure}
    \begin{subfigure}[b]{0.235\textwidth}
        \includegraphics[width=\textwidth]{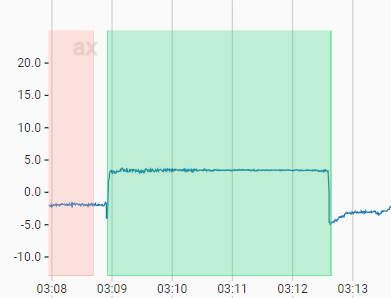}
        \caption{Lying Pattern}
        \label{fig:lying}
    \end{subfigure}
    \begin{subfigure}[b]{0.235\textwidth}
        \includegraphics[width=\textwidth]{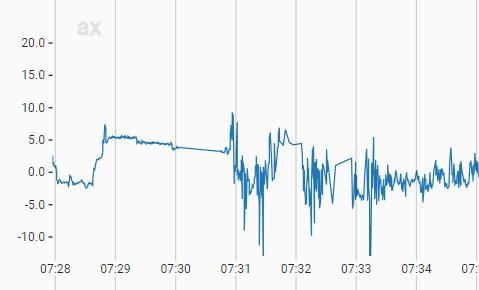}
        \caption{Others activity pattern}
        \label{fig:otha}
    \end{subfigure}
    \caption{Data - Behaviour Patterns}
\end{figure}

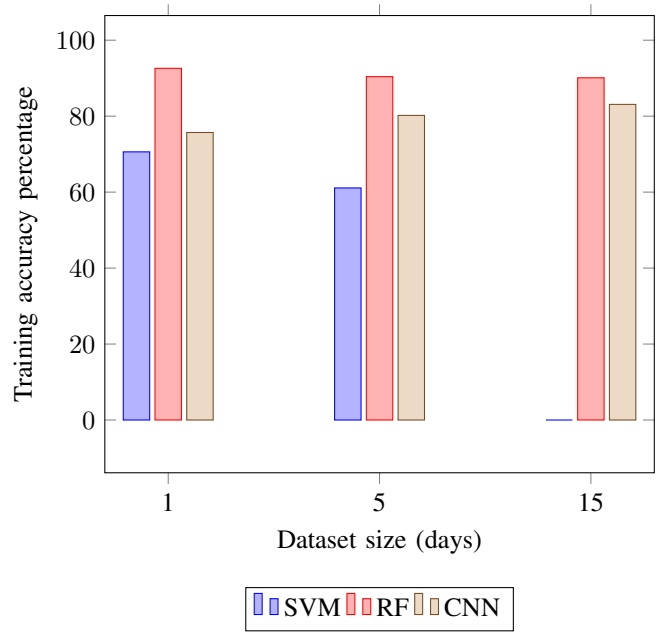
\begin{figure}[h]
\centering
\begin{tikzpicture}
    \begin{axis}[
        width=0.49\textwidth,
        x tick label style={/pgf/number format/1000 sep=},
        symbolic x coords={1,5,15},
        ylabel={Training accuracy percentage},
        xlabel={Dataset size (days)},
        enlargelimits=0.15,
        legend style={at={(0.5,-0.25)},
        anchor=north,legend columns=-1},
        ybar,
        bar width=10pt,
        xtick=data,
        ]
    \addplot 
	coordinates {(1,70.6)(5,61.1)(15,0)};
    \addplot 
	coordinates {(1,92.6)(5,90.4)(15,90.1)};
    \addplot 
	coordinates {(1,75.7)(5,80.2)(15,83.1)};
    \legend{SVM,RF,CNN}
    \end{axis}
\end{tikzpicture}
\caption{Training accuracy of Models vs Dataset Size\textsuperscript{*}}
\label{fig:training_acc}
\end{figure}

\def\thefootnote{*}\footnotetext{SVM training time for 15 days is not included as it is too large.}

Convolutional Neural Networks, particularly one-dimensional CNNs, showed extremely accurate results. They excelled at processing time-series data like accelerometer and gyroscope readings, capturing spatial relationships between time points. CNNs' ability to automatically learn relevant features from raw sensor data over time provided significant advantages in identifying subtle behavioral patterns.\\
All experiments were conducted on a Lenovo IdeaPad Gaming laptop equipped with an AMD Ryzen 5600H processor, NVIDIA RTX 3050 GPU, and 16 GB RAM. 
\subsection{Labelling of data}
Data labeling presented a significant challenge due to the large volume of unlabeled sensor data.
Label Studio was chosen as the data labeling software for its ability to host locally, support time series data labeling, and integrate with ML model training.
To efficiently label the data, patterns for each behavior were identified using CCTV footage available:

Rumination: Typically occurred from 6 pm to 5 am, with acc\_x oscillating around a negative value between -2 and -3, sometimes even beyond -3. \\
Feeding: Showed larger oscillations in acc\_x around 2 to 6, with similar oscillations in other features but varying centers of oscillation. \\
Lying: Appeared as a flat line with tiny spikes representing head movements.\\
Others: All other unlabelled timestamps were labelled as 'Others'. This label lacks any distinct patterns and all cattle movements are included in this section which make it random. Also, since its related to movements, this label is very crucial for the following section of heat detection.

\subsection{Data Preprocessing}
Several crucial preprocessing steps were performed on the raw sensor data, sampled at 2 Hz:

\begin{enumerate}
    \item \textbf{Data Segmentation}: The data was segmented into four categories: feeding, ruminating, lying, and other activities.
    \item \textbf{Noise Removal}: Noise removal techniques were applied to eliminate irrelevant or erroneous data points.
    \item \textbf{Data Normalization}: Data was normalized using the Z-score algorithm:
    \[
    Z = \frac{X - \mu}{\sigma}
    \]
    where \(X\) is the data point, \(\mu\) is the mean, and \(\sigma\) is the standard deviation of the dataset.
    \item \textbf{Rolling Window Function}: A rolling window function with a sliding window of 10 data points was applied.
    \item \textbf{Feature Extraction}:
    \begin{itemize}
        \item \textbf{Statistical Measures}: Calculated minimum, maximum, standard deviation, and average values for accelerometer and gyroscope data in all three axes.
        \item \textbf{FFT}: Utilized Fast Fourier Transform to extract features from rumination and feeding data, which exhibited periodic patterns throughout the day.
        \[
        X_k = \sum_{n=0}^{N-1} x_n e^{-i 2 \pi k n / N}
        \]
        where \(X_k\) is the FFT output, \(x_n\) is the input signal, \(N\) is the total number of samples, \(k\) is the frequency index, and \(i\) is the imaginary unit.
    \end{itemize}
    \item \textbf{Dataset Creation}: Created new datasets with 6 and 24 parameters.
\end{enumerate}

These preprocessing steps ensured that the sensor data was clean, normalized, and rich with features, enabling effective analysis and model training.

\subsection{Implementation}
The evaluation of each algorithm necessitated the use of different libraries due to varying support and capabilities. Support Vector Machines (SVM) were implemented using sci-kit-learn \cite{tf}, while Random Forests were executed using Yggdrasil Decision Forests \cite{ydf}. For Convolutional Neural Networks (CNN), TensorFlow \cite{tf} was used. It's worth noting that sci-kit-learn, primarily designed as an educational framework for machine learning, lacks GPU training support and production environment capabilities. This limitation may have contributed to the extended training times observed for the SVM model. In contrast, both Yggdrasil Decision Forests and TensorFlow offer GPU acceleration and allow for model persistence, enabling their use in production environments.

\subsection{Results} \label{results}
The project achieved varying levels of accuracy in behavior classification:

Multiple versions of ML models were trained with different preprocessing techniques and ML algorithms.
Epochs for Fig \ref{fig:training_acc} were varied until maximum accuracy was reached with the minimum no. of epochs within 1\% of error.
Accuracies varied from 61.1\% to 90.1\% on the full dataset. 
Fifteen days of labeled data, totaling 1843142 data points, were used for training and evaluation.

These results demonstrate the potential of the approach but also highlight areas for improvement. So we re-evaluated the labelled train set and made a highly accurate set this time focused on 10 days of different cows. This led to more consistent results with the accuracy ranging from 93\% to 97\%.

After this prediction, we displayed a summary of the total number of minutes spent in doing one of the four classified activities in the entire day and then, also created an hourly summary of the time spent on the activities. The latter would be extensively used in the following section.

\section{Estrus Detection}
\subsection{LSTM}
Long Short-Term Memory (LSTM) networks \cite{lstm} are a type of recurrent neural network designed to learn long-term dependencies in sequential data. LSTMs are well-suited for estrus detection in cattle, where behavior patterns over time are crucial.
The key feature of LSTM networks is their memory cell, which can maintain information for extended periods. Each LSTM unit contains a cell state and three "gates" (forget, input, and output) that regulate information flow. 
Mathematically, the operations within an LSTM unit can be described by the following equations\cite{lstm}:

1. \textbf{Forget gate}: 
   \[
   f_t = \sigma(W_f \cdot [h_{t-1}, x_t] + b_f)
   \]

2. \textbf{Input gate}: 
   \[
   i_t = \sigma(W_i \cdot [h_{t-1}, x_t] + b_i)
   \]
   \[
   \tilde{C}_t = \tanh(W_C \cdot [h_{t-1}, x_t] + b_C)
   \]

3. \textbf{Cell state update}: 
   \[
   C_t = f_t \cdot C_{t-1} + i_t \cdot \tilde{C}_t
   \]

4. \textbf{Output gate}: 
   \[
   o_t = \sigma(W_o \cdot [h_{t-1}, x_t] + b_o)
   \]
   \[
   h_t = o_t \cdot \tanh(C_t)
   \]

where:
\begin{itemize}
    \item \( \sigma \) is the sigmoid function.
    \item \( \tanh \) is the hyperbolic tangent function.
    \item \( x_t \) is the input at time step \( t \).
    \item \( h_{t-1} \) is the hidden state from the previous time step.
    \item \( C_t \) is the cell state at time step \( t \).
    \item \( W_f, W_i, W_C, W_o \) are the weight matrices.
    \item \( b_f, b_i, b_C, b_o \) are the bias vectors.\\
\end{itemize}

These equations enable the LSTM network to selectively remember or forget information, making it capable of recognizing patterns over extended sequences of data, making it a powerful tool for tasks such as estrus detection in cattle.
\subsection{Implementation}
\begin{figure}
    \centering
    \begin{tikzpicture}[node distance=0.5cm, auto,
    block/.style={rectangle, draw=black, thick, fill=white, text width=3cm, text centered, rounded corners, minimum height=1cm},
    line/.style={draw, thick, -latex'},
    cloud/.style={draw=black, thick, ellipse, fill=white, minimum height=1cm}]

    % Nodes
    \node [cloud] (start) {Start};
    \node [block, below=of start, fill=green!20] (loadtrain) {Load training data};
    % \node [block, right=of loadtrain, fill=green!20] (loadtest) {Load testing data};
    \node [block, below=of loadtrain, fill=orange!20] (processtrain) {Preprocess train data};
    % \node [block, below=of loadtest, fill=orange!20] (processtest) {Preprocess test data};
    \node [block, below=of processtrain, fill=green!20] (combine) {Combine last n days of train data with test data};
    \node [block, below=of combine, fill=orange!20] (normalize) {Normalize combined data with MinMaxScaler};
    \node [block, below=of normalize,fill=blue!20] (createseq) {Create sequences with 3-day lookback};
    \node [block, right=of createseq, fill=blue!20] (loadmodel) {Load pre-trained LSTM model};
    \node [block, above=of loadmodel, fill=blue!20] (predict) {Predict on test sequences};
    \node [block, above=of predict] (mse) {Calculate Mean Squared Error};
    \node [block, above=of mse] (threshold) {Apply threshold to detect anomalies*};
    \node [block, above=of threshold] (map) {Predict heat based on number of anomalies};

    % Connections
    \path [line] (start) -- (loadtrain);
    % \path [line] (start) -| (loadtest);
    \path [line] (loadtrain) -- (processtrain);
    % \path [line] (loadtest) -- (processtest);
    \path [line] (processtrain) -- (combine);
    % \path [line] (processtest) -| (combine);
    \path [line] (combine) -- (normalize);
    \path [line] (normalize) -- (createseq);
    \path [line] (createseq) -- (loadmodel);
    \path [line] (loadmodel) -- (predict);
    \path [line] (predict) -- (mse);
    \path [line] (mse) -- (threshold);
    \path [line] (threshold) -- (map);
\end{tikzpicture}
    \caption{Estrus Detection Algorithm\\}
    TensorFlow, pandas and scikit-learn are denoted by blue,
green,and orange respectively
   
    \label{fig:flowchart_heat}
\end{figure}
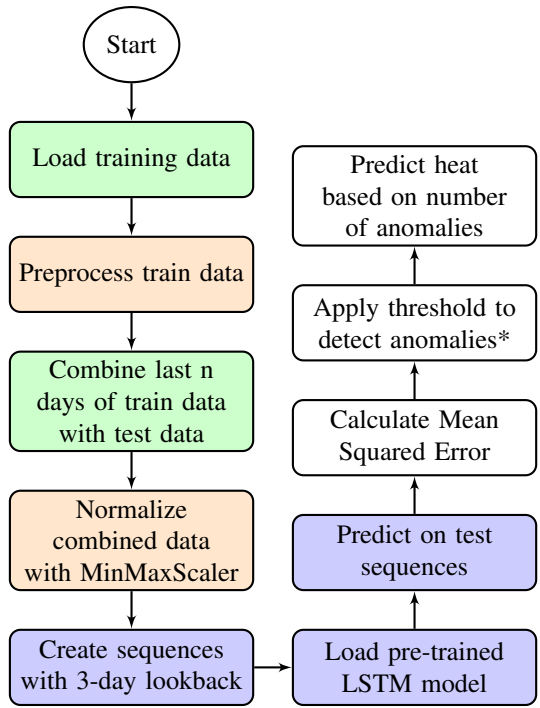

The algorithm was implemented using TensorFlow, pandas \cite{pandas}, and scikit-learn (as illustrated in Fig. \ref{fig:flowchart_heat}. Training was conducted on a 30-day dataset containing the summarised time taken for an activity every hour, with labels provided by the Random Forest classification model described in Section \ref{results}. The LSTM model accepts four input parameters: hour of the day and the duration (in minutes) spent in each of three behavioral patterns (rumination, feeding, and other activities). The model learns these and then a 3 day sequence of previous 3 days' 'others' minutes (based on user time input), is fed into the model. 
It then predicts the time spent in "other" activities. This prediction is compared with a pre-determined, dataset-induced threshold to detect anomalies. If the predicted time is anomalously high, the model flags it as heat.
\subsection{Results}
The threshold values for anomaly detection, and minimum number of anomaly hours required for flagging of estrus was configured based on analyzing heat data from 3 days from different cows. The model was able to accurately predict 4 out of 4 heat days, and flagged 19 out of 20 normal days as 'not heat'. We got only one false positive and the model accuracy on this small
test-set was 96\%.

\section{Disucussion and Conclusion}
% This study demonstrates an efficient approach to cattle behavior detection and estrus prediction using minimal sensor data coupled with advanced machine learning techniques. By utilizing only accelerometer and gyroscope data from a neck collar, our method achieves comparable accuracy to more data-intensive systems. The behavior classification models, particularly Random Forest and CNN, showed promising results in accurately categorizing key cattle behaviors, forming the basis for our LSTM-based estrus detection model. Our approach's key strength lies in its ability to deliver accurate results with reduced hardware requirements, potentially making advanced herd management techniques more accessible to a broader range of farming operations. While the results are encouraging, future work could focus on further refining the models, expanding the dataset, and developing user-friendly interfaces. Ultimately, this research contributes to precision livestock farming by offering an efficient, scalable solution for monitoring cattle health and reproductive status, balancing accuracy with practicality.

This study demonstrates the potential of combining low-cost sensor technology with machine learning techniques for cattle behavior classification and estrus detection. Our approach, utilizing only accelerometer and gyroscope data from a neck collar, achieved promising results comparable to more data-intensive systems.

\subsection{Behavior Classification}

The behavior classification models, particularly Random Forest (RF), showed high accuracy in categorizing key cattle behaviors. The RF model's ability to handle large datasets and provide feature importance rankings proved valuable in identifying critical sensor measurements.

The improvement in accuracy from 61.1-90.1\% to 93-97\% after re-evaluation of the labeled dataset highlights the critical importance of high-quality, accurately labeled data in machine learning applications. This underscores the need for careful data collection and labeling processes in future studies.

\subsection{Estrus Detection}

The LSTM-based estrus detection model, built upon the behavior classification results, showed perfect accuracy in our limited test set. While these results are encouraging, the small sample size (4 heat days, 20 normal days) necessitates caution in interpreting these findings. Further testing with a larger and more diverse dataset is crucial to validate the model's performance and generalizability.

\subsection{Practical Implications}

Our approach's key strength lies in its ability to deliver accurate results with reduced hardware requirements. By utilizing only accelerometer and gyroscope data, we've developed a system that could potentially make advanced herd management techniques more accessible to a broader range of farming operations, including smaller farms with limited resources.

The hourly summary of cattle activities provides valuable insights for farmers, allowing for more precise monitoring of herd behavior. This could lead to earlier detection of health issues or changes in reproductive status, ultimately improving herd management and productivity.

\subsection{Limitations and Future Work}

Despite the promising results, several limitations should be addressed in future research:

\begin{enumerate}
    \item \textbf{Sample Size:} The limited sample size, particularly for estrus detection, necessitates further validation with larger datasets from diverse cattle populations.
    \item \textbf{Environmental Factors:} The current study did not account for environmental factors such as temperature, humidity, or seasonal variations, which could influence cattle behavior and estrus patterns.
    \item \textbf{Individual Variation:} While our models showed good overall performance, they may not account for individual variations in cattle behavior, based on their breed and species. Future work could explore personalized models that adapt to individual animals.
\end{enumerate}

Future research directions could include:

\begin{enumerate}
    \item Incorporating additional sensor types (e.g., temperature) to enhance behavior classification and estrus detection accuracy.
    \item Developing more sophisticated anomaly detection algorithms to reduce false positives in estrus detection.
    \item Exploring the integration of this system with other farm management tools to create comprehensive precision livestock farming solutions.
\end{enumerate}

By addressing these limitations and exploring these future directions, we can further refine and improve the effectiveness of our approach in cattle behavior monitoring and estrus detection. This ongoing work has the potential to significantly contribute to the field of precision livestock farming, offering increasingly accurate and accessible solutions for farmers across various operational scales.

% \section*{Acknowledgment}

% The preferred spelling of the word ``acknowledgment'' in America is without 
% an ``e'' after the ``g''. Avoid the stilted expression ``one of us (R. B. 
% G.) thanks $\ldots$''. Instead, try ``R. B. G. thanks$\ldots$''. Put sponsor 
% acknowledgments in the unnumbered footnote on the first page.

\bibliographystyle{IEEEtran}

\balance

\end{document}